# KUDO Interpreter Assist: Automated Real-time Support for Remote Interpretation

Claudio Fantinuoli
KUDO/Mainz University

Giulia Marchesini
KUDO

David Landan
KUDO

Lukas Horak
KUDO

## Abstract

High-quality human interpretation requires linguistic and factual preparation as well as the ability to retrieve information in real-time. This situation becomes particularly relevant in the context of remote simultaneous interpreting (RSI) where time-to-event may be short, posing new challenges to professional interpreters and their commitment to delivering high-quality services. In order to mitigate these challenges, we present Interpreter Assist, a computer-assisted interpreting tool specifically designed for the integration in RSI scenarios. Interpreter Assist comprises two main feature sets: an automatic glossary creation tool and a real-time suggestion system. In this paper, we describe the overall design of our tool, its integration into the typical RSI workflow, and the results achieved on benchmark tests both in terms of quality and relevance of glossary creation as well as in precision and recall of the real-time suggestion feature.

## 1 Introduction

In recent years, the interest in computer-assisted interpreting tools (CAI), especially in the domain of simultaneous interpretation, has significantly increased. CAI tools can be defined as desktop programs or mobile applications specifically designed to assist interpreters in at least one of the several sub-processes of interpretation, for example knowledge acquisition and management, lexicographic memorization, terminology access, and so forth. The tools available to date greatly differ both in the set of functionalities offered and in their architectures. They can be as simple as terminology management spreadsheets available on the user's computer or complex applications deployed on the cloud, integrating basic features to support the daily work of interpreters or advanced functions that aim to automatize them (e.g. Fantinuoli 2018).

CAI tools are relevant to support interpreters in their effort to maintain or increase the quality of their service in a changing professional landscape. This is particularly crucial in light of the recent rise of remote simultaneous interpretation (RSI), and the need to streamline processes, compensate for shorter time-to-events, and meet the constantly high demands in terms of interpretation quality. The transition from an analogue to a digital workspace, i.e. from a physical booth with a hardware console to the immateriality of an artificial environment, has favoured the integration of CAI tools into RSI consoles, increasing the willingness of practitioners to adopt novel technologies that can help them meet these new challenges. Recent advancements in AI-based applications, such as machine translation, speech recognition, and language modelling, have now created the technical conditions to design and integrate novel supportive features inside RSI workflows.

In this paper we present KUDO Interpreter Assist, a computer-assisted interpreting tool specifically designed for the integration in RSI scenarios. Interpreter Assist comprises two main features: an automatic glossary creation tool and a real-time suggestion system. The goal of Interpreter Assist is to shorten the overall preparation time and to increase the precision of the rendition in highly specialized events.

The reminder of this paper is organized as follows. Section 2 introduces the related work in the area of CAI tools and the empirical experiments conducted around them so far. Section 3 describes the general architecture of the tool. Section 4 introduces the evaluation framework, the results of our tests, and the limitations of our evaluation methodology. Finally, section 5 concludes the paper and presents the outlook.

## 2  Related work

Computer-assisted interpreting tools have been proposed by several researchers in the past 20 years or so (Will 2015; Fantinuoli 2012; Rütten 2017; Stoll 2009). Much attention has been devoted to the activities which are peculiar for the pre-event phase, i.e. the time prior to a conference that interpreters spend with the linguistic and extra-linguistic preparation. In this phase, several automated or semi-automated approaches have been proposed, for example the collection of relevant documentation (Fantinuoli 2017a), the extraction of specialized terminology (Fantinuoli 2006) and the exploration of corpora to investigate a new topic (Xu 2018, Fantinuoli 2017b).

A specific strain of research has studied CAI tools as a means for retrieving specialized information while interpreting. Initially, this was done by manually searching terms in a database and presenting results in an intuitive a distraction-free user interface. More recently, however, automatic speech recognition has been proposed as a technology to automate the query system of CAI tools. This automation may solve the shortcomings of traditional tools (e.g. Prandi 2017, Fantinuoli 2017c, Hansen-Schirra 2012), namely the excessive cognitive load required to use them, and extend the scope of information that can be retrieved by the machine (e.g Wang and Wang, 2019, Mellinger 2019). Not only terminology can be retrieved and presented to the interpreter in real time, but also other so called problem triggers, such as numbers and proper names (Fantinuoli 2017c, Rodriguez et al. 2021). Lately, researchers have started investigating the use of machine learning approaches to the automatic prediction of difficult parts of the speech as a means to improve the usability of the tools (Vogler et al. 2019).

Over the years, a handful of empirical studies have been carried out to test the feasibility of the human-machine interaction in the simultaneous modality. From a user perspective, they have mainly focused on the effectiveness of ASR-support during the interpretation of numbers (Desmet et al. 2018; Defrancq and Fantinuoli 2020; Pisani and Fantinuoli 2021), one of the problem triggers of simultaneous interpreting identified in literature (Braun and Clerici 1996; Gile 2009; Setton and Dawrant 2016). In order to measure the impact on the quality of the rendition, these studies have used either mock-up systems (Desmet et al 2018; Canali 2019) or real-life tools (Defrancq and Fantinuoli 2020; Pisani and Fantinuoli 2021). Results seem to suggest that the use of automatic retrieval system leads to an increased precision in the rendition of the problem triggers under investigation. Little is still known, however, about the influence of such tools on the overall performance of the interpreter, a topic that will need to be addressed in future research.

## 3  Architecture

### 3.1  General workflow

Interpreter Assist comprises two main functionalities: an automatic glossary creation and a real-time suggestion feature for terminology, numbers, and proper names. From the user perspective, the typical workflow for using this tool comprises the following phases:

- Creation of a project with identification of the event topic
- Optional assessment of the automatic generated multilingual resources

- Optional sharing the resources with the team members
- RSI Interpretation with real-time suggestions

From the machine perspective, the process comprises the following steps:

- Creation of a mono or multilingual domain-specific corpus
- Extraction of the monolingual terminology from the corpus
- Translation of the terminology in the target language(s)
- Fine-tuning of the baseline models with the generated resources
- Optional fine-tuning with client-generated information
- Real-time automatic access of the resources in the interpreter console

### 3.2 Glossary generation

The first step of the pipeline is the corpus collection. The event specific corpus is obtained with one of the following three methods:

- A set of documents in one or more languages that are relevant for a specific meeting. These so-called preparation documents, generally provided by the event organizers, may comprise minuting of previous meetings, presentations, etc.
- A number of seed words describing a specific topic, for example "renewable energy", "solar panel" and "photovoltaic". Relevant documents are found by means of a general-purpose search engine, downloaded and added to the specialized corpus (cf. Baroni and Bernardini 2004, Fantinuoli 2006).
- A URL of interest, for example https://www.who.int/health-topics/coronavirus, used as a seed to crawl data from a given webpage. The text from the page is scraped and cleaned (removing headers, footers, meta) and saved. Links in the page are also extracted and followed in a breadth-first manner to a specified depth until either there are no links left to crawl, or until the scraped text exceeds a given size. Preference is given to links with the same prefix and domain as the seed URL.

We generated term lists and scores using several approaches to terminology extraction (Ahmad et al. 1999; Bordea et al. 2013; Lefferts 1995; Kozakov et al., 2004; Sclano and Velardi, 2007) on sample data that was similar to the data an interpreter might use in preparing for an event. After comparing the different results, we decided to retain the approach that offered the best compromise between quality of the term ranking and computational cost of running the algorithm. We choose to apply a modified version of the C-value algorithm (Frantzi et al. 2000).

The standard C-value algorithm ignores unigrams, so we introduced a constant that prevents the base-2 logarithm of term length (in words) from being zero for unigrams. For the part of speech (POS) patterns for candidate terms, we start with a standard pattern based on whether the head noun typically appears in initial or final position in a noun phrase, then modify the patterns after an initial analysis of candidate terms. The standard pattern for head-final noun phrases (e.g., English and German) is: ((N | Adj)* N) | ((N | Adj)* N PP N), and for head-initial noun phrases (e.g., French, Spanish, Italian) is: (N (N | Adj)*) | (N PP N (N | Adj)*). However, we modified both English and German after reviewing initial results, so that the final POS pattern for English is: (N | Adj)* N; and for German is: (Adj* N) | (N N). We do not de-compound German nouns during processing.

Rather than convert the input text to lower case and lemmatize, we save considerations of case and morphology until all candidate terms have been collected. Thus, if a term appears in the text in both singular and plural form, we sum the counts and apply them to the singular form, while removing the plural from consideration; however, if only the plural form is encountered (e.g., big data), we keep the plural form rather than convert it to singular (big datum).

Finally, we apply some heuristics and remove from consideration the following: terms that contain certain stop words (typically comparative and superlative adjectives), unigrams that are common for the language, and terms that are left- or right-anchored subsets of a longer term. The top N terms and suggested translations (100 by default) are then presented to the interpreter, sorted in descending order of score. Higher scores indicate predicted usefulness of the term to the interpreter. The monolingual list of terms is then translated using machine translation. The languages supported at the moment of writing are Arabic, Chinese, English, French, German, Italian, Portuguese, Spanish, Japanese, and Russian.

The user has now the possibility to review the generated resource, editing single entries or adding new terms. The resource can be shared with other team members.

### 3.3 Automatic suggestion

The automatic suggestion feature shows in real-time several units of interests in the online interpreter console used during a remote interpretation session. Two kinds of information are showed to the interpreter:

- Specialized terms and their translations
- Entities such as numbers (with unit of measurement) and proper names

This feature is based on three main components concatenated in a cascading system. In the first step, an automatic speech recognition (ASR) transcribes the speech in real-time. The ASR can be fine-tuned by using the data available for the project, namely the glossary itself, or information added by the client, such as a list of participants, product names and the like. This operation allows to increase the precision of the tool especially for out-of-domain words and proper names.

The transcription is sent to a language model (LM) for the identification of the units of interest: while the terminology is matched using the generated and edited multilanguage glossary (or a set of glossaries), the entities, i.e. the numerals and the proper names, are recognized directly by the language model on the unfolding transcription using NER. We decided to adopt a different approach for terms and entities following interpreters' recommendations. In a survey they expressed the desire to be in full control of the terminology and the translations displayed by the tool. While the chosen approach allows interpreters to trust blindly the suggestions offered by the tool, it has the shortcoming of not offering help on terms that have not been curated by the user. In future, a combination of the two approaches for specialized terminology may be introduced.

Once the results are retrieved in real-time, they are sent to the interpreter console of the RSI platform. In order to limit the abundance of information of the console, the UI is designed to display suggestions in a non-intrusive way.

## 4 System evaluation

### 4.1 Dataset

In order to measure the performance of the tool we created two datasets, one for the evaluation of the terminology extraction and the other for the assessment of the automated suggestion feature. For the purposes of the present paper, both tests have been carried out for English as a source language and Italian, Spanish and French as target languages.

To evaluate terminology extraction, we automatically generate three glossaries in three specialized domains: economics, industrial engineering, and medicine. All terminology lists were limited to 100 terms and were extracted from randomly selected .pdf files, which represents a typical format for preparatory documents in the context of interpretation.

As for the automated suggestions feature, we created two datasets: the first comprises nine .wav audio files with general speeches, mostly in the topics of politics, finance, and economy, but also of more generic topics and informal language (three audio files); the second dataset contains two longer .wav files comprising speeches with a higher density of specialized terminology. All files are public speeches and are representative of the speeches interpreted on the platform.

For every .wav file, two glossaries based on size were created: the first is a medium size topic-related glossary of 200 terms; the second is a big size glossary, made of around 10,000 terms. They represent two extremes in terms of glossary size and allow us to test our terminology retrieval approach with different data conditions, assessing the impact of size of the underlaying data structure on the tool usability, especially in terms of a possible deterioration caused by the retrieval of too many unwanted results (false positives).

### 4.2 Evaluation methodology

To test the terminology extraction, we asked three professional interpreters to evaluate the English terminology list created by the tool. Evaluators had to assign each term to one of the following categories: specialized term, general term, or error. We define as "specialized" a term that is not necessarily understandable by people outside of the field, regardless of domain, while "general" indicates generic terms that most people can understand even when they pertain to a specific domain. "Error" marks incomplete or lexically invalid elements. We consider the "specialized" terms to be highly relevant for the creation of an event-glossary.

Using the same three glossaries, we then asked six interpreters (three for the language pair English>French and three for English>Italian) to evaluate the quality of the translation, marking each pair of extracted term and automatic translation as "acceptable" or "unacceptable" for the creation of a professional glossary to be used in an interpretation assignment.

As for the automated suggestions, we annotated each .wav file and prepared a list of all expected results in two categories: glossary terms and entities (numbers and proper names). We then ran the tool with each speech and glossary (medium and large), generating an output made of three elements: transcription of the speech, recognized entities, and terminology. We measured both precision and recall of the results, identifying issues caused by errors in the transcription and measuring the impact of false positives.

As far as named entities are concerned, results can fall into four categories: a) pass, when expected and correct; b) fail ASR, when not found because of an error with the transcription; c) fail REC, when correctly transcribed but not recognized by the LM; and d) false positives, when a result shows up among the results but is not present in the original speech.

As far as terms are concerned, results can fall into six categories: a) pass; b) pass (different spelling), when the system correctly identifies variations in spelling of a same term; c) fail (different spelling), when the system fails to recognize a term because of a different spelling; d) fail (ASR); e) fail (term not matched), when the exact term is correctly transcribed but not recognized by the LM; f) fail (lemma not matched), when the expected term is different from the lemma in the glossary and it was not recognized; and g) false positives.

### 4.3 Results

### 5.3.3 Evaluation of glossary generation

The evaluation of terminology extraction gives us an indication of how the list of automatically generated terms is perceived by interpreters. Across the three domains, the average categorisation of terms indicates that the distribution between specialized and general terms is similar, where a slightly higher number of majority of terms is considered specialized (Table 1).

*Table 1: term categorization*

| Specialized Terms | 52.5% |
|---|---|
| General Terms | 46% |
| Non-terms | 1.5% |

Terms perceived as errors represent around 1.5% of the total. This provides us with a generally positive evaluation of the tool, with very limited errors and a tendency to select specialized terms for glossaries.

*Table 2: rater agreement*

| Total Agreement | 58.67% |
|---|---|
| Partial Agreement | 41.33% |
| Disagreement | 0.00% |

An important value to better put the results of terminology extraction into perspective is the agreement among interpreters. This shows that total agreement is found in less than 60% of the cases, with partial agreement representing the remaining results. With "partial agreement" we refer to cases in which two out of three interpreters agreed on classifying a term as specialized, general, or error. Total disagreement, which would see evaluators marking a term as three different categories, was never found in our 300 examples. Total agreement seems to be higher for specialized terms while partial agreement for general terms. This indicates a high variability in the way final users evaluate the level of specialisation of single terms, probably depending on personal experience, past subject knowledge, etc. While our limited data only allows us to set a hypothesis, further experiments are needed to shed light on this.

*Table 3: quality of term translations*

| Translations | EN>FR | EN>IT |
|---|---|---|
| Correct | 91.2% | 89.4% |
| Incorrect | 8.8% | 10.6% |

Translation quality was especially well received by interpreters. Our first attempt at evaluating translations from English to French and to Italian saw a very high percentage of terms considered correct for a professional interpreter glossary (Table 3).

### 5.3.3 Evaluation of automatic suggestions

For the automated suggestions we calculate precision and recall values for both entity recognition and glossary terms using the audio corpus introduced before. Table 4 reports the average value among the nine .wav files with the more general speeches. Both precision and recall are computed based on glossary size (M=200, L=10.000). For term recognition a bigger glossary implies a higher number of false positives and, as a consequence, a decreasing precision (from 98.99% to 88.82%). Recall values are relatively low on this general dataset. Typical errors in glossary term recognition are mostly related to a) 5-grams and short 1-gram terms (e.g. "right", "fee") and b) verbs, when their form does not correspond to the lemma (e.g. "addressed" for "address", "studying" for "study").

*Table 4: general corpus without fine-tuning*

|  | Precision | Recall | F1 |
|---|---|---|---|
| Entities | 89.83% | 90.61% | 89.89% |
| Terms (M) | 98.99% | 77.53% | 86.58% |
| Terms (L) | 88.82% | 77.53% | 82.49% |

Our second step addressed ASR fine-tuning as a means to improve the quality of the transcription and consequently of the retrieved results. We fine-tuned the models with a list of words – terms and proper names – simulating the availability of such information in the event organisation pipeline. The results in Table 5 are promising. While not fixing the entirety of the issues, ASR fine-tuning allows us to gain some percentage points in all categories, with a spike of 4.92 % increase in the recall value of glossary terms.

*Table 5: general corpus with fine-tuning*

|  | Precision | Recall | F1 |
|---|---|---|---|
| Entities | 90.03% | 93.20% | 91.39% |
| Terms (M) | 99.05% | 82.45% | 89.59% |
| Terms (L) | 89.26% | 82.45% | 85.46% |

Values in Table 5 represent an average. The tool is generally able to maintain a high percentage of precision and recall in most of the test scenarios, and the highest results achieved show promise. Named entity recognition in particular showed peaks of 100% correct results for all parameters (precision, recall, and F1). Glossary term recognition is less precise: F1 for glossary (M) reaches a best value of 97.78% (with P=100%, R=95.65%) and for glossary (L) a percentage of 96.55% (with P=100%, R=93,33%). Lowest values achieved are still quite high percentages: 84% for named entities and 81% for glossary terms. There was only one notable exception, a generic speech about social issues, which scored a 76.19% in glossary (M) and a 68.90% in glossary (L) with a higher number of errors and false positives impacting results. These preliminary data show that, while the tool tends to remain stable regardless of text or domain, end results can fluctuate depending on the terms we want to recognize and the glossary chosen as reference.

*Table 6: specialized corpus without fine-tuning*

|  | Precision | Recall | F1 |
|---|---|---|---|
| Entities | 86.40% | 92.68% | 89.42% |
| Terms (M) | 100.00% | 96.30% | 98.11% |
| Terms (L) | 93.33% | 96.30% | 94.67% |

Another factor we considered in honing results, in this case exclusively related to glossary terms, is the specificity of the lexicon. Since the glossaries used by interpreters tend to be highly specialized, we performed a second test with two longer .wav audio files which presented a higher lexical density and more specific terminology. The results are presented in Table 6. Both precision and recall reach a better quality compared to the general language corpus, especially in terms of recall. On this dataset we once again applied ASR fine-tuning to improve the transcription quality. The results of this operation are presented in Table 7. They show a slight improvement from the baseline without fine-tuning. However, the effect of fine-tuning seems to be less prominent with respect to the more general corpus.

*Table 7: specialized corpus with fine-tuning*

|  | Precision | Recall | F1 |
|---|---|---|---|
| Entities | 86.63% | 94.19% | 90.23% |
| Terms (M) | 100.00% | 96.30% | 98.11% |
| Terms (L) | 93.33% | 96.30% | 94.67% |

In the analysis of the named entity recognition we paid specific attention to the category of numerals, which is particularly interesting in a real use case scenario. Our first test considered 95 numerals, and among them 93 (around 97%) were immediately recognized, and 2 ran into ASR errors that we were able to correct with ASR fine-tuning, reaching a pass rate of 100%. Our second test yielded slightly worse but still very good results, with a pass rate of 33/35 (around 94%). In this case 2 numerals encountered a recognition error, and because ASR fine-tuning was not aiming at this particular entity type, no gains could be achieved.

An important parameter related to the automated suggestions is the system latency, i.e. how long it takes to process a term, recognize it, and display it on screen. We conducted another specific test taking into account the same nine .wav audio files we used for our first test, and selected three terms for each, both general and specific and in various lengths (x-grams). Measuring latency in seconds for every one of these terms, we measured a response time that varies from a minimum of 1.1 seconds to a maximum of 2.3 seconds, with an average value of 1.6 seconds. This is within the typical ear-voice-span of interpreters and should be considered acceptable for the target use.

### 4.4 Limitations

The empirical results reported herein should be considered in the light of some limitations. Our evaluation was performed with a rather small and standardized benchmark test. For this reason, no generalisations on the performance of the tool can be done. Variables like domain, speech features, accents, to name just a few, may have a huge impact on the performance of the tool and on its usability.

In the specific case of terminology extraction, agreement among interpreters would deserve more attention, for example to determine thresholds of acceptability. An important limit to take into consideration is how these results, being evaluated by humans, are inherently subjective and depend on the evaluator's perspective and background, as indicated by a low interrater agreement on term relevance. Input file type could be expanded (e.g. with URLs), and a differentiation between domain-specific and non-domain-specific terms could be introduced.

Finally, the test of automated suggestions only examines the situation of a specific moment in time: it could become iterative to consistently monitor changes in performance in terms of both precision and recall, and it would benefit from additional data.

## 5  Conclusion

In this paper we presented Interpreter Assist, a novel tool designed for simultaneous interpreters working in a remote setting. The tool offers support with the automatic creation of multilingual glossaries as well as with real-time suggestions of terms, numbers and proper names during the interpretation session. We built a dataset to benchmark the performance of the tool in real-life conditions, with typical speeches, topics and interpreters' glossaries. Preliminary results are encouraging.

While the relevance of the terms automatically extracted by the tool varies greatly among different evaluators, posing a considerable challenge in the tool's ability to meet different user's needs and expectations, the quality of the automatic translated terms is high. As far as the real-time suggestion feature is concerned, with a F1 value of around 98%, real-time suggestions perform well for specialized terminology and numerals, both with and without fine-tuning. However, recall values on general speeches still need to be improved. Performance for named entities increases considerably with fine-tuning. This supports the idea that an event-based pipeline to fine-tune the engines with event-related information is paramount to achieve high quality results. To successfully integrate this in the overall architecture of the tool, we aim at combining the automatic fine-tuning of the models by means of the data obtained during the glossary preparation pipeline with a human-in-the-loop step which will allow users to add ad-hoc information related to the specific event.

## Reference

Ahmad, Khurshid, Lee Gillam, and Lena Tostevin. 1999. Weirdness indexing for logical document extrapolation and retrieval (wilder). In *TREC*, pages 1–8.

Baroni, Marco, and Silvia Bernardini. 2004. "BootCaT: Bootstrapping Corpora and Terms from the Web." In *Proceedings of the Forth International Conference on Language Resources and Evaluation*. Paris: ELRA.

Bordea, Georgeta, Paul Buitelaar, and Tamara Polajnar. 2013. Domain-independent term extraction through domain modelling. In *The 10th international conference on terminology and artificial intelligence (TIA 2013)*, Paris, France.

Braun, Sabine, and Andrea Clarici. 1996. "Inaccuracy for Numerals in Simultaneous Interpretation: Neurolinguistic and Neuropsychological Perspectives." *The Interpreters' Newsletter* 7: 85–102.

Canali, Sara. 2019. "Technologie Und Zahlen Beim Simultandolmetschen: Utilizzo Del Riconoscimento Vocale Come Supporto Durante l'interpretazione Simultanea Dei Numeri." Università degli studi internazionali di Roma,.

Defrancq, Bart, and Claudio Fantinuoli. 2020. "Automatic Speech Recognition in the Booth: Assessment of System Performance, Interpreters' Performances and Interactions in the Context of Numbers." *Target. International Journal of Translation Studies*.

Desmet, Bart, Mieke Vandierendonck, and Bart Defrancq. 2018. "Simultaneous Interpretation of Numbers and the Impact of Technological Support." In *Interpreting and Technology, Language Science Press*, 13–27. Language Science Press.

Evans, David A. and Robert G. Lefferts. 1995. Clarittrec experiments. In *Information processing & management*, 31(3):385–395.

Fantinuoli, Claudio. 2018. "Computer-Assisted Interpreting: Challenges and Future Perspectives." In *Trends in E-Tools and Resources for Translators and Interpreters*, edited by Isabel Durán Muñoz and Gloria Corpas Pastor. Leiden: Brill.

Fantinuoli, Claudio. 2017a. "Computer-Assisted Preparation in Conference Interpreting." *Translation & Interpreting* 9, no. 2: 24–37.

Fantinuoli, Claudio. 2017b"Computerlinguistik in Der Dolmetschpraxis Unter Besonderer Berücksichtigung Der Korpusanalyse." Edited by Oliver Čulo, Slivia Hansen-Schirra, and Stella Neumann. *Annotation, Exploitation and Evaluation of Parallel Corpora*: 111–46.

Fantinuoli, Claudio. 2017c. "Speech Recognition in the Interpreter Workstation." In *Proceedings of the Translating and the Computer 39*. London: Proceedings of the Translating and the Computer 39.

Fantinuoli, Claudio. 2012. "InterpretBank - Design and Implementation of a Terminology and Knowledge Management Software for Conference Interpreters." PhD thesis, University of Mainz.

Fantinuoli, Claudio. 2006. "Specialized Corpora from the Web for Simultaneous Interpreters." In *Wacky! Working Papers on the Web as Corpus.*, edited by Marco Baroni and Silvia Bernardini, 173–90. Bologna: GEDIT.

Frantzi, Katerina, Sophia Ananiadou, and Hideki Mima. 2000. Automatic recognition of multi-word terms: the c-value/nc-value method. In *International journal on digital libraries*, 3(2):115–130.

Gile, Daniel. 2009. *Basic Concepts and Models for Interpreter and Translator Training: Revised Edition*. 2nd ed. Amsterdam: John Benjamins Publishing Company.

Hansen-Schirra, Silvia. 2012. "Nutzbarkeit von Sprachtechnologien Für Die Translation." *Trans-Kom* 5, no. 2: 211–26.

Kozakov, Lev, Youngja Park, T Fin, Youssef Drissi, Yurdaer Doganata, and Thomas Cofino. 2004. Glossary extraction and utilization in the information search and delivery system for ibm technical support. In *IBM Systems Journal*, 43(3): 546–563.

Mellinger, Christopher. "Computer-Assisted Interpreting Technologies and Interpreter Cognition: A Product and Process-Oriented Perspective." *Revista Tradumàtica. Tecnologies de La Traducció* 17 (2019): 33–44.

Pisani, Elisabetta and Claudio Fantinuoli. "Measuring the Impact of Automatic Speech Recognition on Interpreter's Performances in Simultaneous Interpreting." In *Empirical Studies of Translation and Interpreting: The Post-Structuralist Apporach*, edited by Wang Caiwen and Zheng Binghan. Routledge, 2021.

Prandi, Bianca. 2017. "Designing a Multimethod Study on the Use of CAI Tools during Simultaneous Interpreting." In *Proceedings of the 39th Conference Translating and the Computer*, 76–88. Geneva: Tradulex.

Rodriguez, Susana, Roberto Gretter, Marco Matassoni, Daniele Falavigna, Àlvaro Alonso, Oscar Corcho, and Mariano Rico. 2021. "SmarTerp: A CAI System to Support Simultaneous Interpreters in Real-Time." In *Proceedings of Triton 2021*.

Francesco Sclano and Paola Velardi. 2007. Termextractor: a web application to learn the shared terminology of emergent web communities. In *Enterprise Interoperability II*, 287–290. Springer.

Rütten, Anja. 2007. *Informations- Und Wissensmanagement Im Konferenzdolmetschen*. Frankfurt am Main: Peter Lang.

Setton, Robin, and Andrew Dawrant. 2016. *Conference Interpreting: A Complete Course*. Benjamins Translation Library (BTL), volume 120. Amsterdam & Philadelphia: John Benjamins Publishing Company.

Stoll, Christoph. 2009. *Jenseits Simultanfähiger Terminologiesysteme*. Trier: Wvt Wissenschaftlicher Verlag.

Vogler, Nikolai, Craig Stewart, and Graham Neubig. 2021. "Lost in Interpretation: Predicting Untranslated Terminology in Simultaneous Interpretation." *ArXiv:1904.00930 [Cs]*.

Wang, Xinyu, and Caiwen Wang. "Can Computer-Assisted Interpreting Tools Assist Interpreting?" *Transletters. International Journal of Translation and Interpreting* 3 (2019): 109–39.

Will, Martin. 2009. *Dolmetschorientierte Terminologiearbeit. Modell Und Methode*. Gunter Narr Verlag.

Xu, Ran. 2018. "Corpus-Based Terminological Preparation for Simultaneous Interpreting." *Interpreting* 20, no. 1: 29–58.